\documentclass[ds,nolinenumbering]{mscthesis}

\begin{document}
\pagestyle{plain}
\fixemptypage
\setcounter{page}{0}
\title{Time series classification of satellite data using LSTM networks: an approach for predicting leaf-fall to minimize railroad traffic disruption}

\author{Hein de Wilde$^{1}$, Ali Mohammed Mansoor Alsahag$^{1}$,Pierre Blanchet$^{2}$}
\affiliation{%
  \institution{$^{1}$Informatics Institute, University of Amsterdam}
  \streetaddress{1098XH Science Park}
  \city{Amsterdam}
  \country{The Netherlands}\\
  \institution{$^{2}$Meteory}
}
\email{hein.dewilde@student.uva.nl, a.m.m.alsahag@uva.nl, pierre.blanchet@meteory.eu}

\begin{abstract}
Railroad traffic disruption as a result of leaf-fall cost the UK rail industry over £300 million per year and measures to mitigate such disruptions are employed on a large scale, with 1.67 million kilometers of track being treated in the UK in 2021 alone. Therefore, the ability to anticipate the timing of leaf-fall would offer substantial benefits for rail network operators, enabling the efficient scheduling of such mitigation measures. However, current methodologies for predicting leaf-fall exhibit considerable limitations in terms of scalability and reliability. This study endeavors to devise a prediction system that leverages specialized prediction methods and the latest satellite data sources to generate both scalable and reliable insights into leaf-fall timings. An LSTM network trained on ground-truth leaf-falling data combined with multispectral and meteorological satellite data demonstrated a root-mean-square error of 6.32 days for predicting the start of leaf-fall and 9.31 days for predicting the end of leaf-fall. The model, which improves upon previous work on the topic, offers promising opportunities for the optimization of leaf mitigation measures in the railway industry and the improvement of our understanding of complex ecological systems.
\end{abstract}

\maketitle

\section{Introduction}
\label{sec:introduction}
Railroad transportation represents an essential artery for both passenger travel and freight movement. Ensuring its safety and efficiency is paramount, particularly as demands for rapid and frequent services intensify. However, external elements have the potential to disrupt rail operations significantly, with weather conditions being among the key perpetrators \cite{lin2020quantitative}. One weather-related disruption that particularly influences railroad operations during the autumn season is the falling of leaves on railroad tracks. As rail networks often traverse through densely vegetated areas, the seasonal cycle, particularly the advent of autumn coupled with evolving weather conditions, heightens the risk of leaves falling onto the rail tracks. Upon contact with the rails, these leaves, especially under damp conditions, form a layer of low adhesion on the tracks \cite{vasic2008laboratory}, significantly hampering the effective acceleration and braking of trains. Traction between wheel and rail can decrease to a level ten times lower than required \cite{Watts_2021} and the stopping distance can increase by a factor of two to three \cite{Leaf_fall_video}, leading to extended journey durations and potentially rail accidents. Additionally, electrical insulation due to adhesive layers of leaves covering the railroad can disrupt track circuitry, further exacerbating delays due to difficulties in accurately determining train locations on the tracks \cite{networkrail}.

With thousands of tonnes of leaves descending on railroads each autumn \cite{thameslink}, the consequences spiral into hazardous and expensive scenarios for both commuters and rail network operators. Delays as a result of leaf-fall cost the UK rail industry over £300 million per year. The necessary mitigation measures, such as deploying leaf-busting trains and descaling machines \cite{networkrail}\cite{thameslink}, are employed on a large scale, with 1.67 million kilometers of track being treated in the UK in 2021 alone \cite{lbtrain}. The ability to anticipate the timing of leaf-fall would offer substantial benefits for rail network operators, enabling efficient scheduling of such mitigation measures. Regrettably, current methodologies for predicting leaf-fall exhibit considerable limitations. Ground observation-based studies, while precise, are infeasible due to the impracticality of building a scalable, ground-based monitoring system. To exemplify, a commonly applied technique for ground-based vegetation monitoring is the utilization of “phenocams”, which are devices that monitor the foliage of one or several trees by capturing time-lapse imagery \cite{NSF_NEON}. However, the installation of such devices across expansive railroad networks would be a costly and tedious operation. As an alternative, many studies utilize satellite data to monitor large-scale vegetated areas. However, while scalable, these studies often employ outdated data sources and ineffective prediction methods, rendering them unreliable. To illustrate, over the last few years, the resolution of many satellite data types has dramatically improved from hundreds of meters to just several meters per pixel, enabling vegetation monitoring with much higher precision. This underlines the inaccuracy of models using legacy satellite data and highlights the opportunity cost of failing to work with the state-of-the-art.

In light of these limitations, this study endeavors to devise a prediction system that leverages specialized prediction methods and the latest satellite data sources to generate both reliable and scalable insights into leaf-fall timings. By implementing machine learning methods that are designed to handle long-term dependencies within time series data, the aim is to accurately capture the complex dynamics inherent to phenological events like leaf-fall. A dataset is constructed that constitutes globally available satellite data on a wide range of different features that are expected to influence leaf-fall dynamics. It includes weather variables like temperature and precipitation, as well as satellite-derived indexes that measure vegetation health and canopy density. This data is then combined with ground-truth data on leaf-fall events to train the prediction model to identify leaf-falling days within the time series data. The aim is to lay the foundation for a location-agnostic prediction system that can preemptively tackle leaf-induced rail disruptions, therefore reducing delays and accidents, curtailing operational costs, and enhancing travel safety and efficiency. 

This study presents a predictive framework that integrates advanced satellite remote sensing data with machine learning techniques to forecast leaf-fall events along railway networks. The approach begins with the identification of key environmental variables and satellite-derived vegetation indices, such as temperature, precipitation, and canopy health, that are most relevant to leaf-fall dynamics. These features are used to train machine learning models capable of capturing complex temporal patterns in multivariate time series data. By combining satellite-derived predictors with ground-truth observations, the proposed system is designed to detect the timing of leaf-fall events with high reliability. The framework is evaluated for its predictive accuracy, scalability, and practical applicability to operational railway management. Through this approach, the study aims to enable more efficient planning of mitigation measures, reduce delays and disruptions caused by leaf contamination, and contribute to safer and more reliable rail transport during the autumn season.

\section{Literature review}
\label{sec:related_work}
Before modeling complex vegetation dynamics like leaf phenology, an initial understanding is required of what data features are expected to contribute to the accurate prediction of leaf-fall. Literature on the topic notes that plant health involves a large number of environmental and physiological processes, such as meteorological factors (light, photoperiod, temperature, precipitation, humidity, wind, and gasses), edaphic factors (topography, slope, exposure, and soil properties), as well as biotic factors (pests, diseases, and competition) \cite{menzel2002phenology}. Across literature, temperature is noted as one of the key factors influencing phenology dynamics \cite{anderson2005factors}\cite{wolf2017flowering}\cite{menzel2002phenology}. However, the exact relationship between simple meteorological factors and leaf-fall is faint. Research dating back to the mid-20th century noted that plants respond to many meteorological and environmental factors, yet the best method to analyze impacts on plants would be to “ask the plants themselves” \cite{schnelle1955pflanzen}. The researchers hereby referred to plant instruments that monitor vegetation dynamics within the plants themselves. Recent research acknowledges the use of such observation tools for improving phenology modeling, but suggests future studies should primarily focus on using new tools that can scale observations from the species-level to the landscape-level \cite{piao2019plant}. In this context, the utilization of satellite data emerges as a pivotal approach.

\subsection{Satellite-enabled leaf-fall prediction}
Satellite technologies offer an essential tool in the endeavor to accurately predict the timing of leaf-fall. They provide broad-scale, timely, and precise data that can be leveraged to monitor and assess vegetation dynamics and weather conditions, which, as explained above, are key factors influencing leaf-fall.

The utility of satellite data in understanding vegetation dynamics is multifaceted. To illustrate, satellites equipped with multispectral imaging capabilities, such as the Sentinel-2 satellite operated by the European Space Agency (ESA), capture data across spectral bands on different wavelengths, ranging from ultra-blue to short-wave infrared light \cite{sentinel}. By deciphering this spectral information, insights can be gained into various vegetation attributes that can signal impending leaf-fall. One common method to examine vegetation health is the manipulation of spectral bands to generate indexes that offer specific insights into vegetation status. Examples include the Normalized Difference Vegetation Index (NDVI), which provides a quantifiable measure of vegetation greenness \cite{weier2000measuring}, and the Normalized Difference Moisture Index (NDMI), which is a measure of water stress in vegetation \cite{ndmi}. Both indexes can serve as vital indicators for leaf-fall prediction, with declining NDVI and NDMI values potentially signaling the onset of leaf-fall.

Moreover, the Sentinel-1 satellite, also operated by ESA, offers additional complementary data in the form of radar measurements \cite{sinergise}. Radar data can be utilized to generate the Radar Vegetation Index (RVI), providing a measure of vegetation cover that can supplement the insights obtained from multispectral data \cite{szigarski2018analysis}. The value of radar data is particularly apparent in differentiating between varying vegetation heights and densities. This ability can be important for leaf-fall prediction, as it facilitates the distinction between tree canopies, where leaf-fall events occur, and understory vegetation such as shrubs and grass.

However, it is important to note that while the satellite data described above offers invaluable insights, it forms just one component of a comprehensive leaf-fall prediction system. The integration of ground-based observations, meteorological data, and advanced predictive algorithms further enhances the predictive accuracy and reliability. Collectively, this multifaceted approach harnesses the potential of both established and emergent technologies to offer a robust, scalable, and efficient solution to predicting leaf-fall events, thereby enabling timely and effective response strategies for minimizing rail disruptions due to leaf-fall.

\subsection{Applications beyond railroad management}
While the railroad use case is the primary incentive for conducting the study, this research endeavor signifies a substantial contribution to a wider range of scientific fields. Its relevance expands beyond merely enhancing the operational efficacy of railroad management, with potential implications specifically for climate-related research and ecological studies. Firstly, by enabling accurate predictions of leaf-fall timings, by extension, the duration of the growing season can be estimated. As the length of a tree's growing season correlates with its carbon uptake \cite{zani2020increased}\cite{calinger2023century}, the ability to predict the growing season duration with precision could provide invaluable insights for modeling carbon cycles — a crucial focus of current ecological and environmental research. Secondly, the tools and methods developed through this study could set a precedent for other remote sensing applications. Advancements in the field of satellite-based phenology monitoring could foster similar predictive modeling endeavors in related sectors, such as agriculture \cite{khanal2020remote}, forestry \cite{guimaraes2020forestry}, and biodiversity conservation \cite{reddy2021remote}. Ultimately, the broader application of these methods and insights may contribute to a better understanding of the underlying climate phenomena and their impacts, enabling the development of effective climate change strategies.

In summary, this study has scientific relevance that extends beyond its immediate application. It leverages a multifaceted approach to address an operational challenge while also providing potentially impactful insights for environmental research, climate change studies, and ecological modeling. The following section explores existing work that closely relates to this research endeavor and highlights the shortcomings of existing work that this study aims to improve upon.

\subsection{Related work}
Donnelly et al. (2018) \cite{donnelly2018autumn} have compared in situ observations on autumn leaf phenology to satellite data on the same phenomenon. The aim of the study was to assess the effectiveness of satellite remote sensing in capturing autumn leaf phenology as determined by in situ observations. Their research, which examined data over several decades, revealed discrepancies between the two data collection methods. The authors conclude that these discrepancies can be attributed to the satellites’ low-resolution signal, which struggled to capture the heterogeneity of multifaceted landscapes. To overcome these challenges, the authors propose that future efforts could focus on enhancing the satellite data through increased spatial and temporal resolutions. This suggestion serves as an important point of departure for this study, which intends to build upon the latest satellite advancements and leverage them to predict leaf-fall patterns.

The issue of capturing the complexity and variety inherent in landscapes presents an ongoing challenge for satellite remote sensing. Natural, land-based environments typically consist of diverse vegetation species and varied canopy structures, which can differ markedly across both spatial and temporal scales. Thus, accurately capturing these dynamic aspects of vegetation using satellite data can be a complex task. Fortunately, the spatial and temporal resolution of satellite data has seen significant improvements in recent years. Modern satellites can capture data with spatial resolution as detailed as 10 meters – a substantial advancement from the 5 km$^{2}$ resolution utilized in the study by Donnelly et al. Such high-resolution data allows for a more granular representation of landscapes and enables the precise monitoring of leaf phenology and other vegetation dynamics, which is a factor that is critical for this study's leaf-fall prediction model. Additionally, the increasing number of satellites orbiting the Earth have reduced satellite revisit times, allowing for more frequent monitoring of the same area. Consequently, the temporal resolution of satellite data is enhanced, leading to more up-to-date observations and timely tracking of phenological changes, which is another critical aspect for creating accurate predictions on leaf-fall. 

In essence, the approach presented in this study builds upon the current advancements in satellite technology to address the limitations highlighted by Donnelly et al. (2018). By harnessing the improved resolution data, the aim is to create accurate predictions of leaf-fall patterns across complex and diverse landscapes, contributing to the broader goal of mitigating disruptions in railroad operations caused by leaf-fall.

Not only the spatial and temporal resolution of the data, but also the machine learning approach to predict phenological events is expected to play an important role. Czernecki, Nowosad \& Jabłońska (2018) \cite{czernecki2018machine} have conducted a study that evaluates different statistical models for predicting phenological phases with the use of satellite and meteorological data. For each phenophase, considering phases for both spring and autumn, the performance of several regression-based machine learning techniques was compared. The model types included in the study were multiple linear regression, lasso, principal component regression, generalized boosted models, and random forest. The authors found that choosing a specific set of predictors and applying robust preprocessing procedures is more important for the final results than the selection of a particular statistical model. They also found that spring phenophases mostly relate to meteorological indices, whereas for autumn phenophases, there is a stronger information signal provided by satellite-derived vegetation metrics. Over all phenophases, the average RMSE for the best models was 6.3 days, while the RMSE per phenophase varied seasonally from 3.5 to 10 days. The RMSE for autumn phenophases was relatively high compared to other seasons, with the RMSE of the best models ranging between 7 and 10 days. 

Building upon the findings of Czernecki, Nowosad \& Jabłońska (2018), this study aims to optimize the prediction of autumn pheno-phases through the employment of Long Short-Term Memory (LSTM) networks, more on which is explained in section 3.4. The use of LSTMs in this study introduces a more advanced machine learning technique for phenological predictions compared to the regression-based models evaluated in the study described above. The goal is to improve upon the reported RMSE values for autumn phenophases by leveraging LSTM's capacity to model the complex, long-term dependencies that are inherent to phenological events like leaf-fall. Furthermore, this study will operationalize the findings of Czernecki, Nowosad \& Jabłońska (2018) regarding the value of satellite-derived vegetation metrics for autumn phenophase prediction. By incorporating these metrics into the LSTM model, the intention is to capture the strong information signal that these indices provide, thereby further improving the precision of the leaf-fall predictions. Also, in line with the authors’ emphasis on robust preprocessing, the goal is to implement meticulous data wrangling and normalization procedures, ensuring the quality and reliability of the input data, and in turn, enhancing the performance of the LSTM model. Through these advancements, this research aims to build on existing work and contribute valuable new insights into the field of phenological modeling.

There are several papers that specifically address the problem of leaves falling on railroads. Patil (2020) \cite{patil2020machine} leverages a deep learning approach to predict the NDVI of vegetated areas around railroads. The predicted NDVI values are then used to identify track segments that will possibly be impacted by leaf-fall, setting the priority level of a track segment to low, medium, or high. However, while this approach marks railroad tracks with a priority level based on the predicted NDVI value, there is no direct indicator for the timing of leaf-fall, leaving railroad operators with a level of uncertainty. Also, while it is known NDVI is indicative of leaf health, there are many more factors that are expected to contribute to an accurate prediction of leaf-fall. Therefore, the belief is that the approach proposed by Patel (2020) is an oversimplification of the situation. This study aims to improve upon the study described above by providing railroad operators with detailed, actionable insights into the moment of leaf-fall. Also, the prediction model will leverage not only NDVI, but a wide range of meteorological and vegetational indicators to accurately model phenological patterns.

Gosenshuis (2020) \cite{gosenshuis2020quantification} developed a model for Dutch railroad operator ProRail that quantifies the amount of leaf biomass near railroad crossings. The model relies not on satellite-derived data, but on parameters inherent to the risk of leaves falling on railroads, like tree height, distance between tree and track, angle of tree to track, and wind direction. The end result is a map that marks the railroad crossings that possess the highest risk of being affected by leaf-fall, based on the predicted amount of leaf biomass on each railroad crossing. While this approach might yield actionable results for ProRail, the prediction model is dependent on highly specific data provided by the railroad operator. Therefore it is not widely applicable and not scalable to other geographical areas. This study aims to overcome that limitation by making use of a prediction model that relies on data that is publicly available on a global scale, therefore making its results applicable by any railroad operator.

\section{Methodology}
\label{sec:methodology}
This section offers a comprehensive description of the methods that were applied throughout the research project. Each stage of the data pipeline is explained in its dedicated subsection, providing detailed insights into the data collection process, as well as the data transformations that were performed to prepare the data for the machine learning stage. Next, the specific machine learning techniques that were applied are explained and a justification is given for the chosen approach. The section concludes with an explanation of the methods used for evaluation of the prediction model and the generalizability of the model is highlighted.

\subsection{Leaf-falling data collection}
There are various phenology organizations and research groups around the world that have been recording phenological events for numerous years. These on-the-ground observations on the moment of leaf-fall can be used as ground-truth data for training and testing the prediction model. Examples of data sources that could provide validation data for this research project are Pan European Phenology DB \cite{PEP725}, Global Phenological Monitoring Programme \cite{GPM}, International Phenological Gardens of Europe \cite{IPG}, USA National Phenology Network \cite{usanpn}, Harvard Forest (United States) \cite{HF}, TEMPO (France) \cite{tempo}, and Nature’s Calendar (United Kingdom) \cite{NC}. The organizations behind these potential data sources were selected based on the availability of data on their website and the focus of their research efforts. 

Based on the completeness and thoroughness of the various data sources, the data provided by the Harvard Forest Data Archive \cite{HFDA} was deemed the most usable for this research project. Researchers at Harvard School of Forestry have been collecting phenology data through ground observations from 1990 until present day, for both spring phenological processes and autumn phenological processes. Specifically, the “lfall” feature of the “hf003-04 fall phenology” dataset \cite{HF003} was of interest to this research project, as it indicates the percentage of leaves that have fallen from a given tree on a given date. Of the data sources that were considered, the Harvard Forest dataset was the most detailed ground-truth data on leaf-fall available, thus, was picked to be used as validation data for the leaf-falling prediction model. As the dataset of interest is openly available in the Harvard Forest Data Archive, it was downloaded from this source in CSV format and stored locally for later analysis.

\subsection{Satellite data collection}
A variety of factors were considered in the selection of satellite datasets, including the coverage area, the data resolution, and the temporal span. To align with the selected validation data from Harvard Forest, the collection area for the satellite data was confined to the same region. According to the Harvard Forest website, all trees included in their phenological research are located within a 1.5-kilometer radius around their headquarters \cite{HF003}. Consequently, an area extending 2.0 kilometers from the Harvard Forest headquarters was selected for data collection. In terms of data resolution, it was essential to select a satellite dataset with a level of precision sufficient for tracking the phenological processes of individual trees. This specificity was needed because the validation data was collected through ground observations, meaning the data was collected on a per-tree basis. Satellite data with a resolution of 100 meters would fail to provide accurate phenological insights for individual trees, considering the vast species diversity in most forests, each with distinct phenological cycles. On the other hand, 10-meter resolution satellite data would provide comparatively precise phenological insights for single trees. Lastly, concerning the data's temporal span, a multi-year dataset was desirable, as this would enable the model to identify and learn the recurring leaf-fall patterns unfolding over the years. Essentially, each year included in the dataset presents an additional learning cycle for the model, with a longer dataset equating to more training data for the prediction model.

\subsubsection{Multispectral satellite data}
\hfill\\
The first satellite data source that was selected to be included in the leaf-falling prediction model is the “Harmonized Sentinel-2 MSI: MultiSpectral Instrument, Level- 1C” dataset. Sentinel-2 is one of the latest satellites in the constellation of the European Space Agency. It provides wide-swath, high-resolution, multispectral imagery with a global 5-day revisit frequency \cite{S2}. The MultiSpectral Instrument (MSI) onboard Sentinel-2 samples various spectral bands: visible and near infrared at 10 meters, red edge and shortwave infrared at 20 meters, and atmospheric bands at 60 meters spatial resolution. The spectral bands can be used to construct remote sensing-derived vegetation indexes, like NDVI, NDMI, and NDWI. These three indexes were eventually incorporated in the prediction model, as they are indicative of the health, greenness, and water content of vegetation \cite{Antognelli_2021}. To illustrate, the figures included in Appendix A show the substantial decrease in NDVI across Harvard Forest during autumn. As the chlorophyll concentration in leaves decreases during autumn \cite{primka2019synchrony}, their ability to absorb visible red light and reflect near-infrared light – which are the light frequencies used to calculate NDVI – also decreases, resulting in a lower NDVI value. Thus, a decrease in NDVI over time signifies the development of phenological processes like leaf-coloring, leaf senescence, and eventually, leaf-fall.

Google Earth Engine (GEE), a multi-petabyte catalog of satellite imagery and geospatial datasets, houses the Sentinel-2 datasets. Before remote sensing imagery is made available on GEE, there are several types of data corrections applied by publishers to account for errors. Three common sources of error for imagery products that are often addressed are atmosphere (i.e. air chemistry), topography (i.e. elevation), and geometry (i.e. pixel consistency) errors \cite{Engelstad_Carver_2021}. For the Sentinel-2 datasets available on GEE, a high level of data correction has already taken place. The task that remains is to make a choice between the Surface Reflectance (SR) and Top-of-Atmosphere Reflectance (TOA) dataset \cite{S2_catalog}. SR data has received the highest level of preprocessing in an attempt to best represent the actual conditions on the ground. However, the Sentinel-2 SR data is available from 2017, whereas the Sentinel-2 TOA data is available from 2015. Therefore, the choice was made to utilize the TOA dataset for this study, as two years of additional training data could prove to be valuable for the prediction model. Also, a high level of data correction has already been applied to the TOA dataset, meaning there is sufficient accuracy and reliability in the data for it to be included in the prediction model. 

For the Harvard Forest area with a radius of 2 kilometers, Sentinel-2 TOA data was extracted from GEE using a Javascript script that specifies the area and time frame of interest. The script extracts the data programmatically by interacting with the GEE API and requesting data from the area and time frame specified in the script. Through a pre-built cloud detection algorithm, the data extraction script ensured only data that was not obstructed by clouds was being extracted. This resulted in missing values throughout the dataset, but these are accounted for during the preprocessing stage. The geospatial Sentinel-2 data was stored in GeoTIFF format, which is a commonly used format to store raster graphics and image information. Google Cloud was chosen as the long-term storage location for the extracted data, as it seamlessly integrates with GEE.

\subsubsection{Meteorological satellite data}
\hfill\\
The second satellite data source that was selected to be included in the leaf-falling prediction model, is the “ERA5-Land Daily Aggregated - ECMWF Climate Reanalysis” dataset. This dataset contains daily data on 50 climate-related variables from 1963 until present day \cite{ERA5}. ERA5 is a reanalysis dataset, meaning it combines past observations with today’s climate models to generate a consistent time series of the climate variables. Reanalyses are among the most-used datasets in the geophysical sciences, as they provide a comprehensive description of the observed climate as it has evolved during recent decades \cite{reanalysis}. From the ERA5 dataset, the features that are most relevant to predicting leaf-fall were picked to be included in the prediction model, including (but not limited to) temperature, precipitation, solar radiation, and volumetric soil water content. 

The ERA5 dataset has a resolution that is much lower than that of Sentinel-2: 11,132 meters. Consequently, the ERA5 dataset is not able to provide precise data for individual trees. However, this is not considered a problem for this particular research project, as weather is not expected to significantly vary within small areas of land. For example, even though tree X and tree Y are a kilometer apart, they experience about the same temperature and surface pressure. For other features, like precipitation, more precision could be desired. However, ERA5 was considered to be the best available option because of the large number of climate features included in the dataset and the highly consistent data on a daily frequency. 

In contrast to the high-resolution Sentinel-2 data that consists of many 'pixels', the low resolution ERA5 data covers the Harvard Forest area with just one 'pixel'. Therefore, it suffices to store the ERA5 data for Harvard Forest in a tabular format like CSV, rather than in a raster format like GeoTIFF. Thus, after extraction from GEE, the ERA5 data was stored in CSV file format on Google Cloud. 

\subsection{Data wrangling}
\subsubsection{Leaf-falling data}
\hfill\\
First, all features were dropped that had nothing to do with this research project. The features that remained were the date, the tree ID, and the percentage of leaf-fall. Next, all pre-2015 data was dropped. The reason for this is that Sentinel-2 data is available from 2015 onwards, since that is the year the Sentinel-2 satellite was launched. Therefore the scope of this research was set to the years 2015 to 2022.

Next, the goal was to create a continuous time series with a daily frequency. For the original dataset, data was collected once every week (with some slight deviations). Also, the data was limited to the autumn, thus only containing data from September to December. To transform the original data into a continuous time series with a daily frequency, all of the days were added for the months January to August and the missing days were added for the months September to December. For the months January to August, a value of zero was used for the percentage of leaf-fall. This is justified, as the original data shows that the leaf-falling process starts from September. For the missing days in September to December, a value of NaN was used for the percentage of leaf-fall. As a next step, these values were filled with numerical values using linear interpolation.

Now the leaf-falling period could be derived from the interpolated leaf-falling values. Naturally, the leaf-falling period starts when the percentage of leaf-fall is more than zero for the first time, and the leaf-falling period ends when the percentage of leaf-fall reaches one hundred. This feature was added to the dataset as a boolean value, yielding True for rows with a leaf-fall value above 0 and below 100, and yielding False for the remaining rows. 

In order to select satellite data for individual points in an area, the coordinates of those points are required. Therefore, before the satellite data could be combined with the leaf-falling data of individual trees, the coordinates of the trees in the leaf-falling dataset needed to be estimated. This was done through a separate dataset \cite{HF003} that was provided by the Harvard Forest Data Archive that indicates the site where each tree is located. Using Google Maps, these sites were found on the map and coordinates were noted for each of the sites. Not all sites were to be found on Google Maps, so the trees for which no coordinates could be determined were dropped from the dataframe.

\subsubsection{Sentinel-2 data}
\hfill\\
First, Python library Rasterio was used to read the raster data in the original GeoTIFF files that were extracted from GEE. Next, the estimated coordinates of the trees were used to extract NDVI, NDWI, and NDMI data for the trees in the Harvard Forest area. This data was then stored in a Pandas DataFrame format and just like with the leaf-falling data, the dataset was converted into a continuous time series with a daily frequency. Unfortunately, due to multi-day revisit times and cloud obstruction, for some months only a few data points were available, though this was solved with relative ease by means of linear interpolation. Appendix B shows an example of NDVI raster data that is partially incomplete due to cloud obstruction.

\subsubsection{ERA5 data}
\hfill\\
The ERA5 dataset required little data wrangling, as the data was already on a daily interval, without any missing values. Also, because one pixel of ERA5 data already covers the entire area of interest, the data could be stored in tabular format, instead of in raster format. The only data wrangling that was performed on the ERA5 dataset was selecting the relevant climate features from the main dataset and renaming and restructuring them so they could be used alongside the leaf-falling dataset.

\subsection{Model selection}
The main objective of this research is to predict the period of leaf-fall by analyzing the relationship between leaf-falling data and various satellite-derived data features, all of which are assembled in time-series format. As leaf-fall is a complex natural process influenced by multiple interconnected factors that evolve over time, the forecasting task inherently becomes a time-series prediction problem. More specifically, the task at hand is a time-series classification problem, as we seek to categorize each day of the time-series into one of two classes: leaf-falling day or non-leaf-falling day.

Therefore, model selection for this task involves several criteria. Firstly, the chosen model must be capable of handling time-series data effectively, recognizing the sequential nature of the data points and the inherent temporal dependencies. Secondly, it should be capable of handling high-dimensional data, given the multitude of satellite-derived features involved in the analysis. Thirdly, the model should be adept at binary classification tasks, as the objective is to classify each day into one of the two classes. Lastly, it is desirable for the model to have the ability to learn complex relationships and patterns from the data, given the intricate and interconnected factors influencing leaf-fall.

For this prediction task, several machine learning models could be considered, such as logistic regression, decision trees, random forests, or support vector machines. However, these models often fail to recognize and learn from long-term dependencies within a sequence, which is a crucial element in our context, where the leaf-fall could be influenced by climatic and environmental patterns evolving over extended periods of time. In contrast to the machine learning models mentioned earlier, other machine learning models, particularly those from the domain of deep learning, have shown promise in effectively handling complex time-series data. More specifically, Long Short-Term Memory (LSTM) networks were designed to handle such tasks.

\subsubsection{Long-Short Term Memory}
\hfill\\
Long Short-Term Memory (LSTM) networks are a type of Recurrent Neural Network (RNN) that were first introduced by Hochreiter and Schmidhuber in 1997 \cite{hochreiter1997long}. While traditional RNNs can process sequential data, they struggle with retaining information from distant past inputs due to the vanishing gradient problem \cite{Medewar}. LSTM networks overcome this issue with their unique architecture that is specifically designed to overcome the vanishing gradient problem. Unlike traditional RNNs, LSTM networks can learn and remember over extended sequence lengths and capture the temporal dependencies within these sequences. In other words, LSTM networks are capable of 'remembering' information from the past and using this memory to inform future predictions. That attribute is essential when modeling natural phenomena like leaf-fall, as the leaf-fall period could be influenced by climatic and environmental patterns that evolve over extended periods of time. Conventional machine learning models often fail to capture these dependencies effectively, but LSTM can handle these long-term dependencies, making it a promising candidate for our task.

In the LSTM architecture, there are three key components known as the input gate, forget gate, and output gate \cite{Srivastava_2023}. These gates regulate the flow of information inside the LSTM unit and enable the network to keep or discard information based on its relevance and significance. In the context of predicting leaf-fall, the input gate decides how much of the current day's climate and satellite data should be stored in the cell state, which represents the "memory" of the LSTM unit. The forget gate determines which parts of the existing cell state remain relevant for the current prediction and should be retained, taking into account the historical climate and satellite data. Finally, the output gate decides what information from the current cell state is relevant for the current output, which in the case of this research project would be whether the day is categorized as a leaf-falling day or not.

\subsubsection{Sliding window}
\hfill\\
To feed the LSTM model with data, a sliding window approach was employed. When using this approach, the size of the sliding window is an important consideration. Essentially, the window size determines how many days of data the model looks at at a time to find and learn patterns in the data. For example, with a window size of 7 days, the model is trained on a sequence of 7 days of data at a time to predict the leaf-fall period for the next day. The window then “slides” forward by one day, taking the next 7 days of data for the next training step, and the process repeats. By combining the long-term dependency handling of the LSTM architecture and the systematic input of data using this sliding window approach, the aim is to create a model that is capable of accurately predicting whether a day is a leaf-falling day based on multi-year time-series data.

The size of the sliding window must be selected based on the nature of the problem, the characteristics of the data, and the temporal dynamics that are believed to influence the outcomes. A window size of 7 days was selected for training and testing the model, as it strikes a balance between capturing meaningful temporal information and avoiding over-complication of the model. This window size allows the model to incorporate the influence of weather changes occurring over a week, which is believed to be a suitable time span for detecting significant patterns influencing leaf-fall. A smaller window size, such as 3 days, might not fully capture the effects of fluctuating weather conditions on leaf-fall, while a larger window size, such as 30 days, could introduce unnecessary complexity into the model, potentially leading to overfitting. Therefore, a window size of 7 days provides a satisfactory compromise, enabling the model to capture a comprehensive set of influences without overwhelming it with excessive detail. This choice, combined with the optimal configuration of other hyperparameters of the model, serves to enhance the model's capability to accurately predict the leaf-fall period. Further discussion on the configuration of the other hyperparameters of the model can be found in section 3.6.

\subsection{Data transformation}
In addition to the data wrangling steps described earlier, further data transformation was required before the data could be fed to the LSTM model. Specifically, the numerical features used for the prediction model were subjected to min-max normalization, rescaling the range of features to fall within the range [0, 1]. Furthermore, using one-hot encoding, the categorical feature "tree\_species" was converted to a binary format (0 or 1). These transformations were implemented in consideration of the inherent structure of the LSTM model.

For the numerical features, scaling is essential due to the LSTM model's use of activation functions such as the hyperbolic tangent (tanh) and sigmoid within its architecture, specifically within the input, forget, and output gates \cite{Srivastava_2023}. These activation functions transform their inputs into specific ranges: -1 to 1 for tanh and 0 to 1 for sigmoid. In LSTM models, where gates regulate the information flow based on the inputs, large-scale or highly varying numerical features can result in the activation functions producing extreme outputs. These extreme outputs can lead to biases in learning, with certain features dominating the learning process while others are neglected. By ensuring all numerical features operate on a similar scale, the model is given a fair opportunity to learn from all available features, enhancing the overall accuracy of the model.

For the categorical feature "tree\_species", one-hot encoding was employed. The LSTM model, like most neural network architectures, operates on numeric data. Categorical data presented as text needs to be transformed into a numerical form to be processed effectively by the model. One-hot encoding achieves this by creating binary (0 or 1) columns for each category of the feature. This technique preserves the distinct information encapsulated by each category without introducing any arbitrary ordinal relationship, which would be the case with simple numerical encoding.

\subsubsection{Feature engineering}
\hfill\\
To accurately predict the period of leaf-fall, an important consideration is the seasonal cycle that most trees adhere to. The timing of leaf-fall varies across species and locations, but is typically strongly influenced by the time of the year. To capture this information effectively, the "week\_of\_year" feature was incorporated as an additional feature in the dataset. It was derived by converting the raw dates in the dataset to their corresponding week of the year.

This form of temporal data augmentation is helpful to encapsulate the inherently cyclical nature of leaf-fall, thereby capturing the repeating patterns of this seasonal behavior. Unlike the representation of a raw date, which is linear and therefore unable to embody cyclical patterns over a year, the "week\_of\_year" feature provides a circular representation where week 52 naturally loops back to week 1. This representation aligns more closely with the annual cyclical behavior observed in leaf-fall patterns.

The choice was made to handle "week\_of\_year" as a numerical feature, rather than a categorical one. Firstly, because one-hot encoding the "week\_of\_year" feature would significantly expand the feature space, introducing more risk for overfitting. Secondly, because handling "week\_of\_year" as a numerical feature serves the purpose of preserving the ordered and sequential nature of the data. If treated as a categorical feature, the model could misinterpret the 52 distinct weeks as unrelated categories without any inherent relationship or order between them. This could potentially obscure the crucial temporal patterns related to the seasonality of leaf-fall. By treating "week\_of\_year" as a numerical feature, the ordered and continuous nature of time is maintained, enabling the model to discern the patterns that might unfold across consecutive weeks and that might be crucial for predicting leaf-fall periods accurately. Given "week\_of\_year" was treated as a numerical feature, it was scaled along with the other numerical features before being used as input for the LSTM model.

\subsection{Hyperparameter tuning}
Before training and testing the model to produce the final predictions for the leaf-falling periods, an important step is to tune the hyperparameters of the LSTM model. These hyperparameters control diverse aspects of the model that make up its architecture and have the potential to significantly influence the performance of the model. Their optimal settings are usually data-dependent, necessitating a thorough tuning process to find the values that yield the best performance for the given task. The tuning process was performed in an automated manner using the Hyperband optimization method \cite{li2017hyperband}, implemented in the Keras Tuner library \cite{hyperband2}. This method is an adaptive resource allocation and early-stopping strategy to quickly converge on a high-performing model \cite{TensorFlow}. More detail on the configuration of the hyperparameter tuning process can be found in Appendix C.

The hyperparameter tuning process yielded an LSTM model architecture that comprises three layers, each characterized by specific properties. The first LSTM layer has 256 units and utilizes a tanh activation function. The second and third LSTM layers are configured with 32 units each, and both employ a relu activation function. Following these three LSTM layers, the model includes a Dense layer with a sigmoid activation function. The use of a sigmoid activation function in the Dense layer serves a specific purpose in this binary classification problem. Given its output ranges between 0 and 1, the sigmoid function is ideal for representing a probability distribution over the two classes: leaf-falling day and non-leaf-falling day. The Dense layer thus condenses the learned information from the previous LSTM layers into a format suitable for the final prediction.

A binary cross entropy loss function is employed to guide the learning process. This loss function quantifies the dissimilarity between the predicted probabilities and the true class labels, helping the model optimize its predictions over the course of training \cite{Ronaghan_2019}. The selected learning rate for the optimizer in the model is 0.001. This low learning rate facilitates a slower, yet more precise learning process, which can contribute to the detection of subtle patterns in the data, thus potentially enhancing the model's predictive accuracy. 

In terms of regularization, a dropout rate of 0.1 is applied in the first LSTM layer. This technique is instrumental in mitigating the risk of overfitting on the training data, which aids in improving the model's generalization performance on unseen data. The second and third LSTM layers do not use dropout, which means that during the training process, all units in these layers are fully engaged, and no regularization via dropout is applied.

The LSTM architecture described above strikes a balance between model complexity and adaptability. It thereby enhances its robustness against overfitting and improves its potential for predicting the leaf-fall period with high accuracy on unseen data.

\subsection{Model evaluation}
The dataset comprising Harvard Forest leaf-falling data, Sentinel-2 multispectral data, and ERA5 weather data was partitioned into three distinct segments – training, validation, and testing – to ensure a robust evaluation of the LSTM model's predictive performance. The temporal nature of the data, spanning daily intervals from 01-01-2015 to 31-12-2022, required a thoughtful data split that respected the chronological order of observations.

Given the sequential dependency in time series data, the data was divided based on temporal boundaries rather than a random split. This partitioning method respects the inherent temporal relationships, allowing the model to learn and validate from a continuous sequence of historical data. The period from 2015 through 2021 was utilized for training, providing the LSTM model with seven years of sequential data to learn from. Subsequently, the year 2022 served as a validation set, enabling an unbiased evaluation of the model's predictive power on unseen data points.

For the purpose of testing the model's performance, a distinct holdout set was formed by isolating data corresponding to a single tree from the main dataset. The primary rationale behind this approach is twofold. First, it presents the model with entirely unseen data, mirroring a real-world scenario where the model would be presented with data from areas not included in the initial training phase. Second, it provides a way to evaluate the model's ability to generalize the learnt patterns to new, disparate data points, rather than merely recalling specific sequences from the training data.

\subsubsection{Evaluation metrics}
\hfill\\
The performance of the model was assessed by means of a classification report, which computes the precision, recall, and F1 score for both classes, as well as a global accuracy score. This report is particularly advantageous as it differentiates between the two classes – leaf-falling day and non-leaf-falling day – offering a granular view of the model’s performance. By focusing on the evaluation metrics of the leaf-falling day class, a good understanding can be obtained of the model’s ability to predict the leaf-falling period accurately. To assess the temporal aspect of the leaf-falling predictions, the difference (number of days) between the predicted first day of leaf-fall and the actual first day of leaf-fall is noted, which is also done for the predicted last day of leaf-fall and the actual last day of leaf-fall. Then, using these differences, a root-mean-square error (RMSE) is computed that serves as a quantifiable metric of the model’s ability to accurately predict the start and end of the leaf-falling period. The performance of the LSTM model is then compared to the performance of the methods described in previous work on phenological modeling.

\subsubsection{Model generalizability}
\hfill\\
The aim of this study is to develop a model that can make accurate leaf-fall predictions irrespective of geographical location. Therefore, the testing methodology described above is designed to provide a rigorous evaluation of the model's generalizability. The ability to predict leaf-fall patterns based on new, unseen data from an unknown tree is expected to be an indicator of the model’s predictive capabilities for new regions.

The adaptability of the leaf-falling prediction model is directly linked to the universal nature of the Sentinel-2 multispectral data and ERA5 weather data, both of which cover extensive geographic areas. The developed model is designed to learn from patterns and relationships that are not solely tied to localized conditions, so that it can be applied to broader leaf-fall dynamics across diverse regions and ecosystems. Thus, as long as the Sentinel-2 and ERA5 data is available, the model should function proficiently, making accurate leaf-falling predictions regardless of the geographical area in question. 

The next sections delve deeper into the practical implementation and evaluation of the leaf-falling prediction model. Its performance on the distinct testing data is scrutinized and insights are provided into the model’s adaptability and generalizability. Also, while moving forward with the described methodology, it is important to consider potential limitations and challenges when applying the prediction model in practice. Therefore, the efficacy of the model in real-world scenarios is explored while acknowledging potential limitations of the proposed approach and considering areas for future refinement and enhancement. 

\section{Results}
\label{sec:results}
This section presents a comprehensive analysis of the performance metrics of the developed prediction system, providing an overview of its proficiency in predicting leaf-fall occurrences. The performance evaluation is divided into various segments, each examining a particular facet of the model's effectiveness. The section concludes with a brief analysis of the real-world practicality of the model, which is then further discussed in the subsequent section.

\subsection{Classification report}
The core evaluation of the model's performance hinges on its ability to accurately classify the occurrence of leaf-falling days, as this metric directly impacts the reliability of the model’s predictions. The classification report (Table 1) lists the precision, recall, and F1 scores for both the leaf-falling day and non-leaf-falling day class. For the leaf-falling day class, the model exhibits a precision of 0.91, which indicates that the model correctly identified 91\% of the leaf-falling days from all the days it labeled as leaf-falling days. The recall stands at 0.82, which suggests that the model was able to capture 82\% of the actual leaf-falling days from the total leaf-falling days. The F1 score, which is a weighted average of precision and recall, has a value of 0.86. These metrics collectively signify that the model performed fairly well in correctly predicting leaf-falling days, but that refinement to the model might still be necessary to affirm its utility in real-world applications. For the non-leaf-falling day class, the model achieved a precision, recall, and F1 score of 0.97, 0.99, and 0.98 respectively. This high performance is explained by the imbalance in the dataset, which naturally contained more non-leaf-falling days than leaf-falling days. Also, the non-leaf-falling days are ‘easier’ to predict, as the training data shows that a large portion of the year (December until August) never experiences leaf-fall. The model proved to have learned this from the training data, as the graphical representation of the predictions (Appendix D) shows that no leaf-falling days were predicted in that portion of the year. 

\begin{table}[H]
\renewcommand{\arraystretch}{1.2}
\centering
\caption{Classification report}
\label{classification_report}
\begin{tabular}{|p{2cm}|c|c|c|c|}
\hline
\textbf{Class} & \textbf{Precision} & \textbf{Recall} & \textbf{F1-Score} & \textbf{Support} \\ \hline
Leaf-fall & 0.91 & 0.82 & 0.86 & 432 \\ \hline
No leaf-fall & 0.97 & 0.99 & 0.98 & 2483 \\ \hline
\textbf{Macro Avg} & 0.94 & 0.90 & 0.92 & 2915 \\ \hline
\textbf{Weighted Avg} & 0.96 & 0.96 & 0.96 & 2915 \\ \hline
\end{tabular}
\end{table}

\subsection{Learning curves}
The learning curve for the model accuracy (Figure 1a) shows that the model achieved similar accuracy scores across the testing and validation sets, illustrating the model's robustness against overfitting. The accuracy for the testing set was around 0.95 for the first epoch and improved to just under 0.97 in the subsequent epochs, showing the model’s capacity to learn from the training data to slightly improve its already high accuracy. On the validation set, the model’s accuracy swings around the 0.96 mark throughout the epochs. The up-and-down movement of the plot suggests that the validation accuracy is not very stable, but given the difference between the high and low values across the epochs is so small (around 0.05), this instability is not deemed problematic. The learning curve for the model loss (Figure 1b) shows that the loss behaved similarly to the accuracy for both sets, further underlining the model's robustness and adaptability.

\begin{figure}[h]
  \centering

  \begin{subfigure}{0.48\columnwidth}
    \centering
    \includegraphics[width=\linewidth]{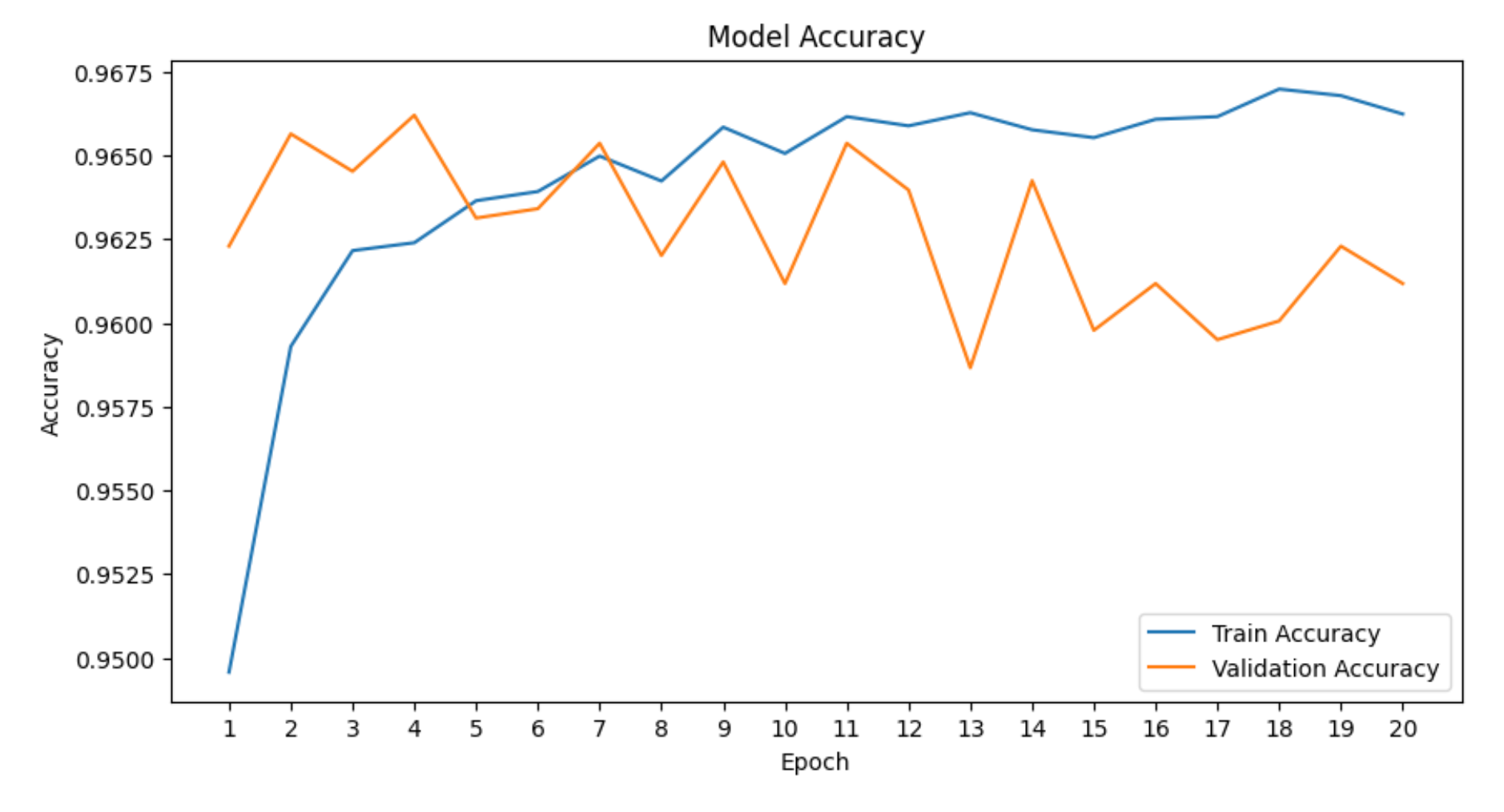}
    \caption{Accuracy per epoch}
    \label{fig:image1}
  \end{subfigure}
  \hfill
  \begin{subfigure}{0.48\columnwidth}
    \centering
    \includegraphics[width=\linewidth]{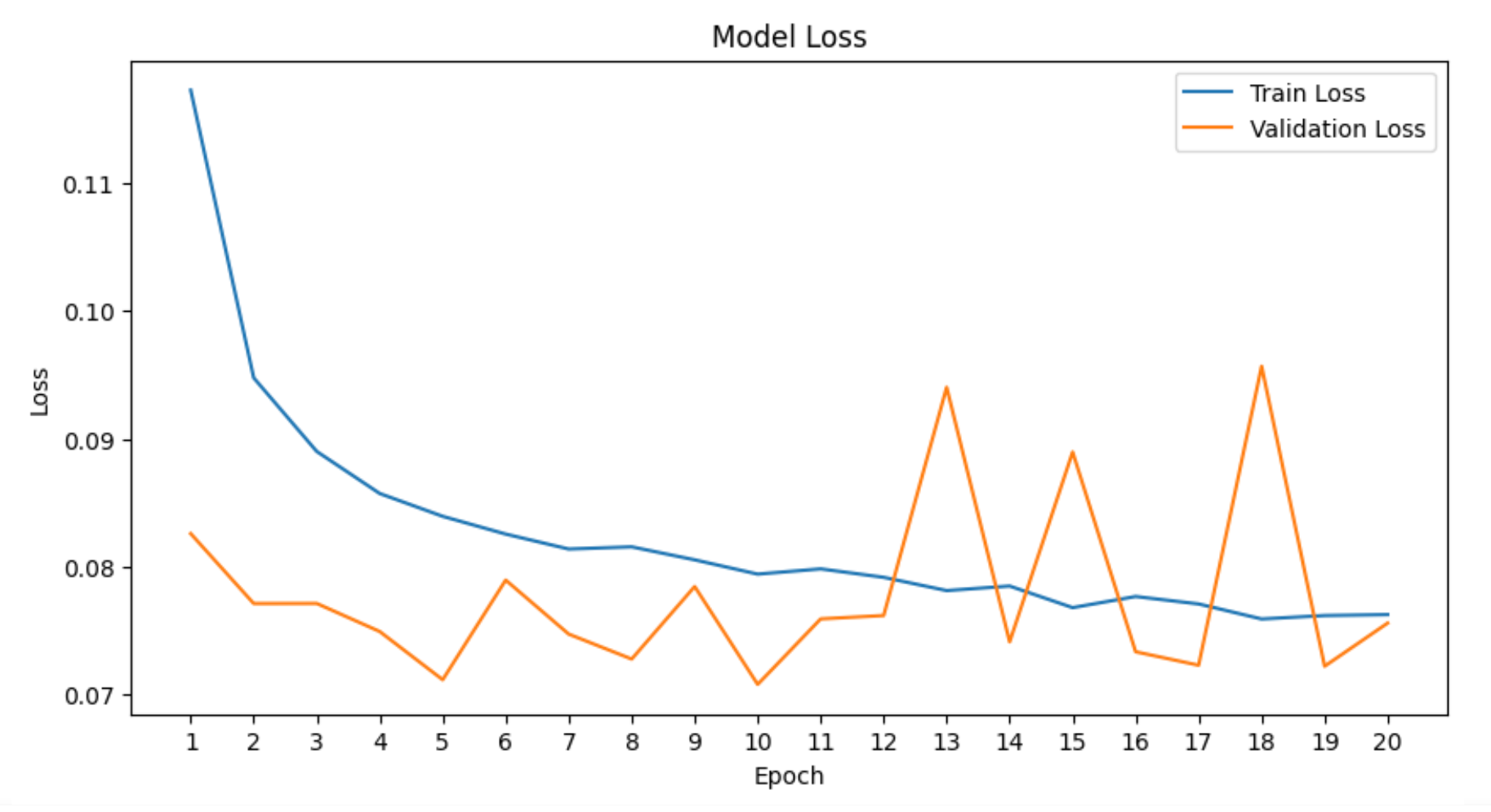}
    \caption{Loss per epoch}
    \label{fig:image2}
  \end{subfigure}

  \caption{Learning curves LSTM model}
  \label{fig:overall}
  \vspace{-10pt}
\end{figure}

\subsection{RMSE}
The results shown in Appendix D offer insight into the temporal aspect of the leaf-falling predictions. Furthermore, Table 2 and Table 3 show the difference between the predicted and actual dates for the start and end of the leaf-falling period. These differences can be used to compute the root-mean-square error (RMSE) of the predictions, similar to how Czernecki, Nowosad \& Jabłońska (2018) \cite{czernecki2018machine} assessed the performance of their regression-based models. The LSTM model described in this study scored an RMSE of 6.32 days for predicting the start of leaf-fall and an RMSE of 9.31 days for predicting the end of leaf-fall, resulting in an overall RMSE of 7.96 days. Given the RMSE of the best-performing models by Czernecki, Nowosad \& Jabłońska (2018) varied between 7 and 10 days (for the autumn phenophases), the LSTM model proposed in this study has proven to be on par with, if not slightly better than, the regression-based models proposed by Czernecki, Nowosad \& Jabłońska (2018).

\newcolumntype{C}[1]{>{\centering\arraybackslash}p{#1}} 
\newcolumntype{Y}{>{\centering\arraybackslash}X} 

\begin{table}[H]
\renewcommand{\arraystretch}{1.2}
\centering
\caption{Predicted vs. Actual first day of leaf-fall}
\label{classification_report}
\begin{tabularx}{\linewidth}{|C{1cm}|Y|Y|C{1.5cm}|}
\hline
\textbf{Year} & \textbf{Predicted} & \textbf{Actual} & \textbf{Difference} \\ \hline
2015 & September 15 & September 17 & 2 days \\ \hline
2016 & September 15 & September 13 & 2 days \\ \hline
2017 & September 10 & September 13 & 3 days \\ \hline
2018 & September 16 & September 25 & 9 days \\ \hline
2019 & September 8 & September 4 & 4 days \\ \hline
2020 & September 8 & September 17 & 9 days \\ \hline
2021 & September 10 & September 21 & 11 days \\ \hline
2022 & September 12 & September 10 & 2 days \\ \hline
\multicolumn{4}{|c|}{\textbf{RMSE:} 6.32 days} \\ \hline
\end{tabularx}
\end{table}

\begin{table}[H]
\renewcommand{\arraystretch}{1.2}
\centering
\caption{Predicted vs. Actual last day of leaf-fall}
\label{classification_report}
\begin{tabularx}{\linewidth}{|C{1cm}|Y|Y|C{1.5cm}|}
\hline
\textbf{Year} & \textbf{Predicted} & \textbf{Actual} & \textbf{Difference} \\ \hline
2015 & November 1 & November 16 & 15 days \\ \hline
2016 & October 31 & November 1 & 1 day \\ \hline
2017 & November 1 & November 9 & 8 days \\ \hline
2018 & November 6 & November 15 & 9 days \\ \hline
2019 & October 26 & November 2 & 7 days \\ \hline
2020 & October 25 & October 25 & 0 days \\ \hline
2021 & November 1 & November 16 & 15 days \\ \hline
2022 & October 27 & November 3 & 7 days \\ \hline
\multicolumn{4}{|c|}{\textbf{RMSE:} 9.31 days} \\ \hline
\end{tabularx}
\end{table}

\subsection{Practical implications}
The results described above indicate that the LSTM model is capable of making temporally accurate leaf-fall predictions, thereby validating the efficacy of the LSTM approach in modeling a complex, time-dependent phenomenon like leaf-fall. The model exhibits strong classification performance and improved temporal predictions as compared to previous work, while showing potential for wide-scale application due to its reliance on globally available data. However, there are still several limitations when it comes to applying the model in practice. Before the model is implemented for real-world decision making, it is important to further examine these limitations, which is what the next section aims to do. The intention is not only to address the constraints and propose solutions for them, but also to provide suggestions for future research directions. By building upon the findings presented in this study, there is significant potential for researchers to further develop academic understanding and practical know-how in the field of phenological modeling. 

\section{Discussion}
\label{sec:discussion}
The first practical limitation of the model developed for this study is that it uses historical data to make predictions on historical dates, while in practice, it is desired to make predictions for future dates in order to efficiently schedule leaf mitigation measures. One way to achieve this goal is to make predictions based on future weather data, which can be obtained through short-term weather forecasting, and future values of satellite-derived indexes like NDVI, which can be obtained by analyzing the trend of these indexes. However, by forecasting data and deriving data through trend analysis, the data inherently carries a level of uncertainty and inaccuracy, possibly affecting the model’s performance. Future research could focus on developing a leaf-fall prediction model that incorporates future weather and satellite data to make predictions on the forthcoming leaf-fall period. Then, after backtesting the model’s performance on observed ground-truth data on leaf-fall, its performance can be compared to the performance of the prediction model proposed in this study.

Another practical limitation of the model is the limited level of detail it provides on the leaf-falling occurrences that it predicts. While the model indicates what days are expected to be leaf-falling days, thus predicting what the start and end of leaf-fall will be, the model does not provide information on the degree of leaf-fall within that period. An analysis of the trajectory of the Harvard Forest leaf-falling process has shown that a large portion of the leaves fall within a short period of time and that the initial and final phase of the leaf-falling period just barely contribute to the total degree of leaf-fall (see Appendix E). Therefore, future research could focus on developing a model that provides more detailed information on the degree of leaf-fall on specific days. This would give railroad operators the actionable insights that are required for effectively scheduling leaf mitigation measures.

Next to the two proposed research directions for enhancing the practicality of the leaf-falling prediction model, there are several ways in which both the performance and reliability of the existing LSTM model could be further improved. First, while the model is trained to recognize general patterns in leaf-falling periods based on weather data and satellite-derived vegetation indexes, not all geographic features that might influence leaf-fall are included in the prediction model. Therefore, the unique intricacies of local ecosystems might affect model performance when the model is applied to an unseen geographical area. For example, while the soil type might be a factor influencing leaf-fall in a certain region, it is not (yet) included as one of the predictive features in the model. To further build upon the robustness of the current LSTM model, future research could focus on expanding the feature space with more data specific to geographic areas, like altitude, soil type, and the potential presence of tree diseases. Also, the quality of the existing weather data could be improved by harnessing novel climate sensing technologies like smart weather balloons \cite{WindBorne_Systems}. Such technologies can provide highly accurate, feature-rich, and localized weather forecasts, which links back to the just discussed desire to utilize the model with forecasted weather data.

In addition to improving weather data quality, the widespread capabilities of satellites could be further embraced by not only utilizing Sentinel-2 multispectral data, but also incorporating Sentinel-1 radar data into the model. This would add a high level of granularity to the dataset, as the Sentinel-1 data could provide valuable information on phenological processes by monitoring the canopy structure of the vegetated area. The reason Sentinel-1 data was not incorporated into this study yet, is because it is highly specialized data that needs extensive preprocessing, adding a significant layer of complexity to the data pipeline. Future research could build upon the work presented in this study to augment the existing data pipeline with radar data from Sentinel-1 and evaluate its effect on the performance of the leaf-falling prediction model.

\section{Conclusion}
\label{sec:conclusion}
This study has demonstrated the utility of LSTM networks in the prediction of leaf-falling patterns based on historical weather data and satellite-derived vegetation indexes. It has been shown that this model is capable of predicting leaf-falling days with a high degree of accuracy, and that it offers potential improvements over traditional methods in terms of temporal accuracy.

Several practical limitations of the model have been acknowledged, notably the use of historical data for predictions and the lack of detail provided on the predicted leaf-falling occurrences. Suggestions for future research include the incorporation of near-term weather forecasts and trend analysis of satellite indexes to enable predictions for future dates, as well as further feature expansion for better accuracy across diverse geographic regions. 

Despite these limitations, the model has shown promise in its generalizability and adaptability, which could potentially enable accurate leaf-falling predictions irrespective of geographical location. The results from this research serve as a starting point for more detailed studies on the application of LSTM networks for the prediction of complex, time-dependent ecological phenomena.

In conclusion, the model developed in this study serves as a proof-of-concept for a novel, machine learning-based approach to leaf-falling prediction. It offers opportunities for the optimization of leaf mitigation measures in the railway industry and the improvement of our understanding of complex ecological systems. The prospect of future research expanding upon this study indicates a promising trajectory for utilizing machine learning to further decode and predict the sophisticated dynamics within our natural environment.

\section{Acknowledgements}
I would like to thank Harvard Forest, the European Space Agency (ESA), and the European Centre for Medium-Range Weather Forecasts (ECMWF) for providing the data used in this study. Additionally, I would like to express my gratitude to dr. Ali M.M. Alsahag of University of Amsterdam for contributing his time and expertise to help improve this manuscript. Lastly, I would like to thank Pierre Blanchet and Alexandre Larroumets of Meteory for providing me with the technical tools and resources that enabled me to conduct this research.

\newpage
\bibliographystyle{plain}
\bibliography{references.bib}

\onecolumn

\appendix
\begin{appendices}

\section{Harvard Forest NDVI data}

\vspace{0.8cm}

\begin{figure}[h]
\centering
\begin{subfigure}{.5\textwidth}
  \centering
  \includegraphics[width=\linewidth]{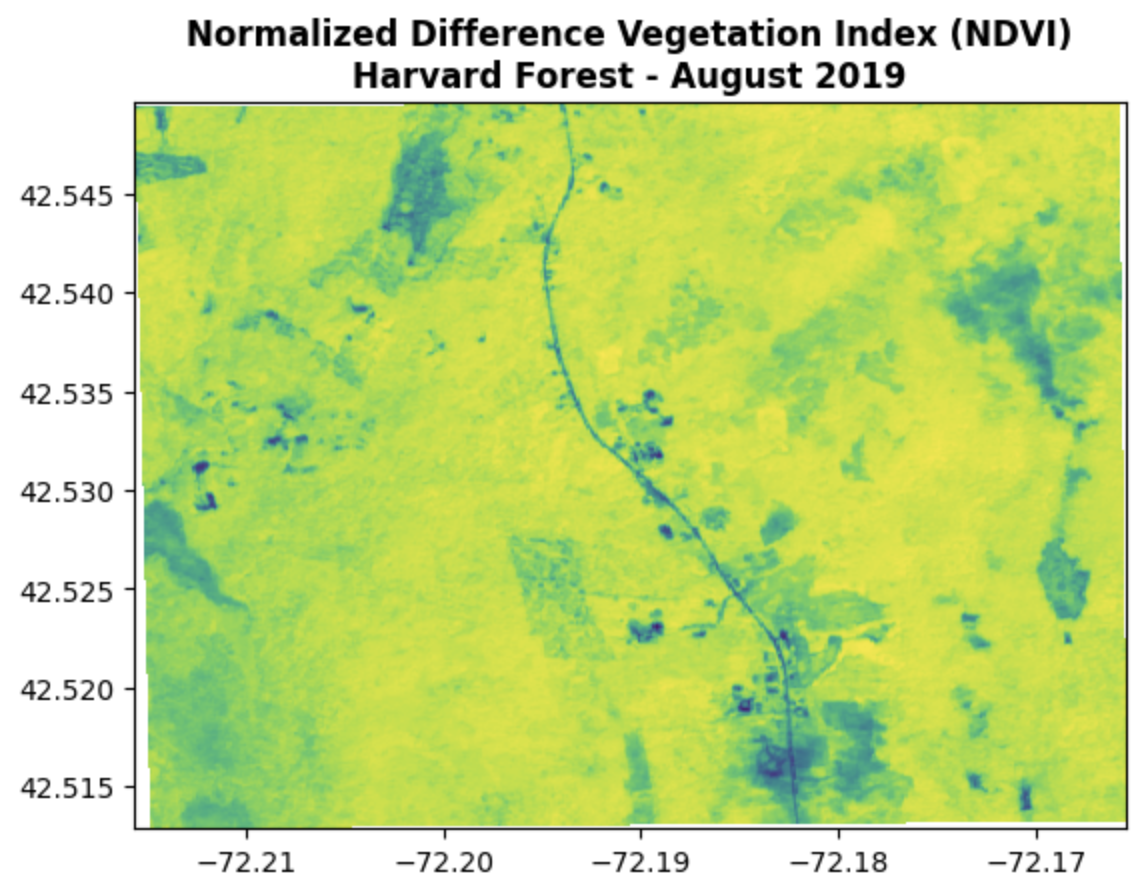}  
  \label{accuracy}
\end{subfigure}
\begin{subfigure}{.5\textwidth}
  \centering
  \includegraphics[width=\linewidth]{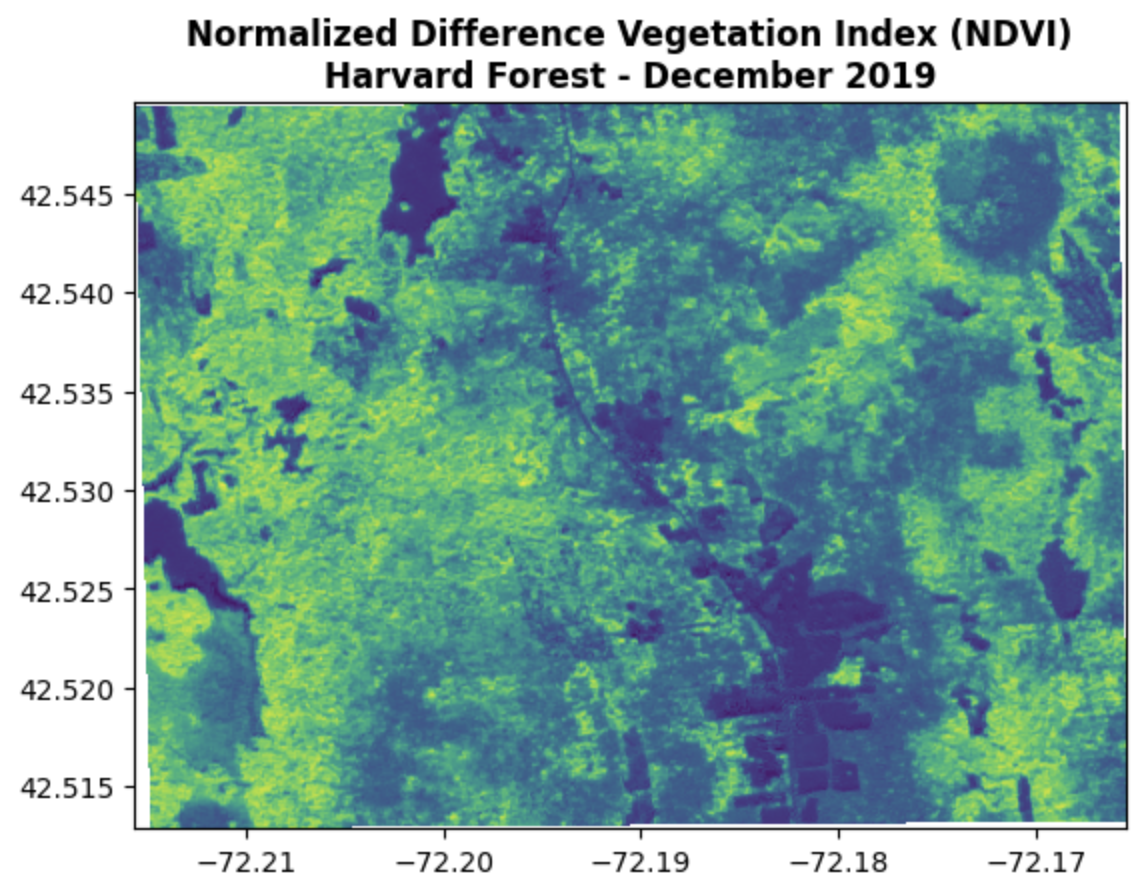}  
  \label{loss}
\end{subfigure}
\caption{Harvard Forest NDVI in August vs. December}
\label{accuracy_loss}
\end{figure}

\begin{figure}[h]
\centering
\begin{subfigure}{.535\textwidth}
  \centering
  \includegraphics[width=\linewidth]{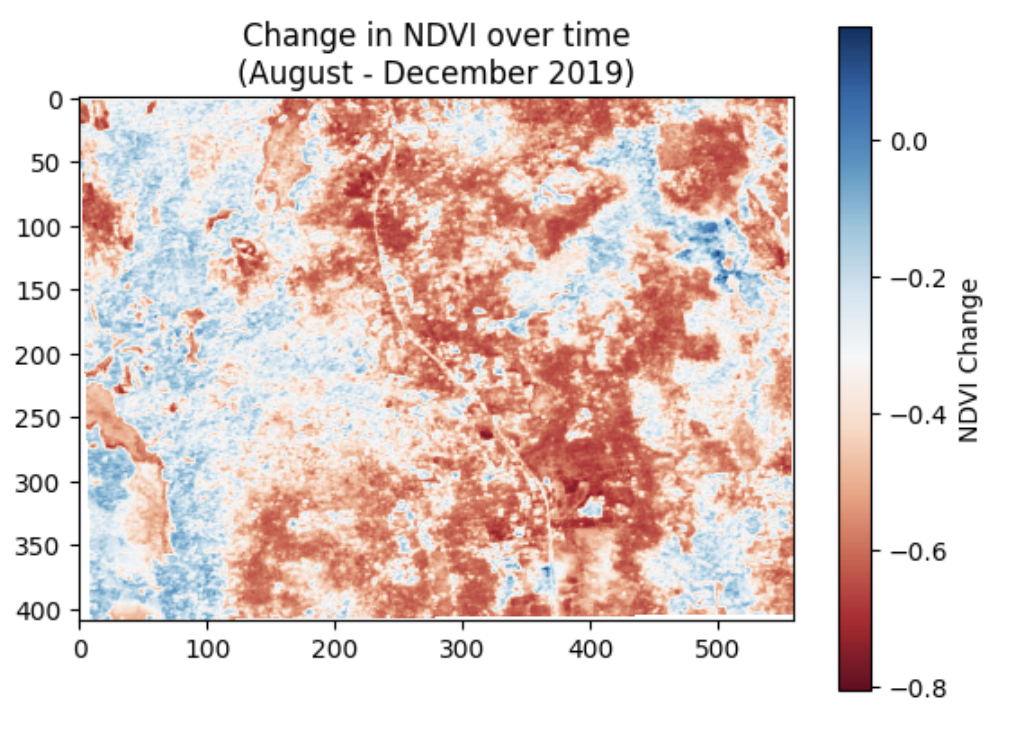}  
  \label{accuracy}
\end{subfigure}
\begin{subfigure}{.465\textwidth}
  \centering
  \includegraphics[width=\linewidth]{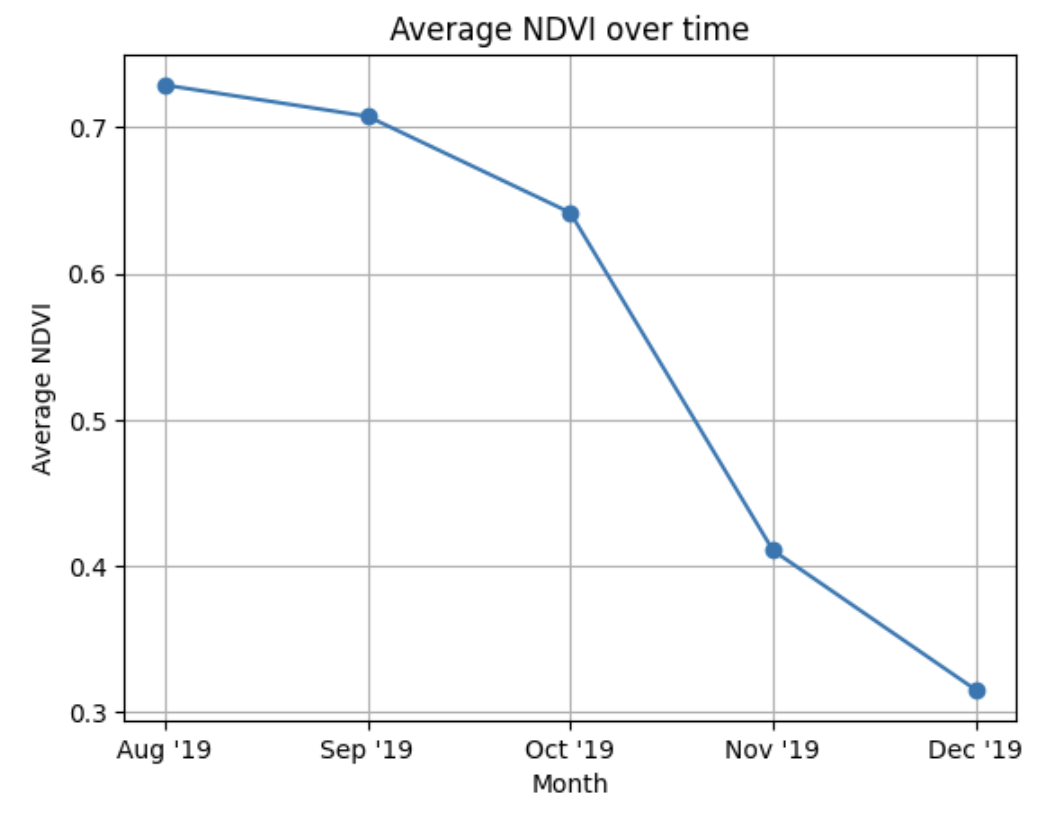}  
  \label{loss}
\end{subfigure}
\caption{Harvard Forest NDVI decrease throughout autumn}
\label{accuracy_loss}
\end{figure}

\newpage

\section{Cloud-obstructed NDVI data}

\begin{figure}[h]
  \centering
  \begin{minipage}{0.6\linewidth}
    \includegraphics[width=\linewidth]{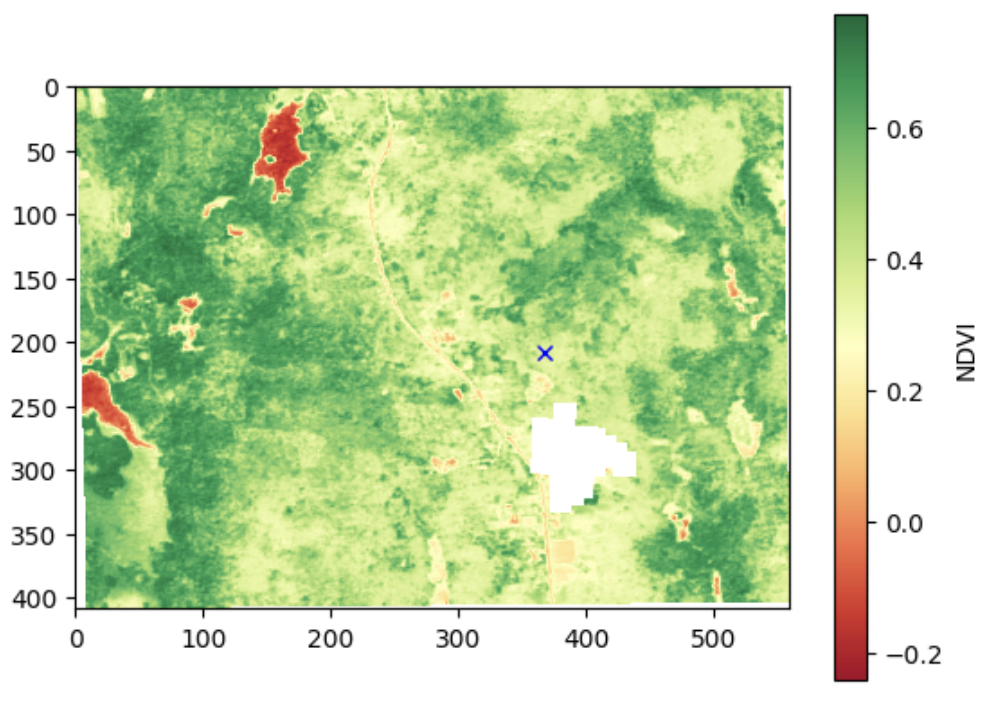}
    \caption{Example of partially cloud-obstructed NDVI data, just south of one of the trees in the leaf-falling dataset (indicated by blue cross)}
    \label{fig:cm}
  \end{minipage}
\end{figure}

\vspace{1cm}

\section{LSTM hyperparameters}
The various LSTM hyperparameters and the defined search spaces that the Hyperband tuner explored, are explained in more detail below. The selected ranges for the hyperparameters were largely based on common practices in machine learning research and on the characteristics of this specific study.

\begin{itemize}
\item \textbf{Number of layers:} This hyperparameter reflects the depth of the LSTM model. A deeper model may theoretically learn more complex patterns, but there is a risk of overfitting if the model becomes excessively deep. Therefore the defined search space ranged from 1 to 3 layers, as more complex models are expected to lead to overfitting, particularly with relatively small datasets such as the one used in this study.
\item \textbf{Number of units in each LSTM layer:} This hyperparameter influences the capacity of the LSTM to capture information in the input sequence. A higher number of units can allow for more complex representations, but may lead to overfitting. The defined search space ranged from 32 to 512 units with a 32 unit interval, in order to explore a broad range of model complexities, from relatively simple to highly complex.
\item \textbf{Activation function:} This hyperparameter controls the non-linearity in the LSTM computations. Different activation functions can model different types of relationships in the data. The defined search space included the relu, tanh, and sigmoid activation functions, due to their prevalent use and proven effectiveness in deep learning applications.
\item \textbf{Learning rate:} This hyperparameter determines the magnitude of the updates made to the model weights at each step of the training process. A higher learning rate may lead to faster convergence, but it can also overshoot the optimal point. Conversely, a lower learning rate may yield a better model but increase the training time. The defined search space included 0.01, 0.001, and 0.0001, as values outside this range are expected to result in unstable learning or slow convergence.
\item \textbf{Dropout rate:} This hyperparameter is a regularization technique to prevent overfitting. It randomly sets a proportion of input units to 0 at each update during training time. The defined search space ranged from 0.0 to 0.5 with a 0.1 interval, in order to examine a variety of regularization strengths.
\end{itemize}

\noindent These chosen hyperparameter ranges allow the LSTM model to explore various configurations, potentially optimizing its ability to capture the leaf-falling patterns without overfitting the data. The LSTM model architecture that the tuner eventually converged on is outlined in section 3.6.

\newpage

\section{Predicted vs. actual leaf-falling days}

\begin{figure}[h]
\centering
\begin{subfigure}{.44\textwidth}
  \centering
  \includegraphics[width=\linewidth]{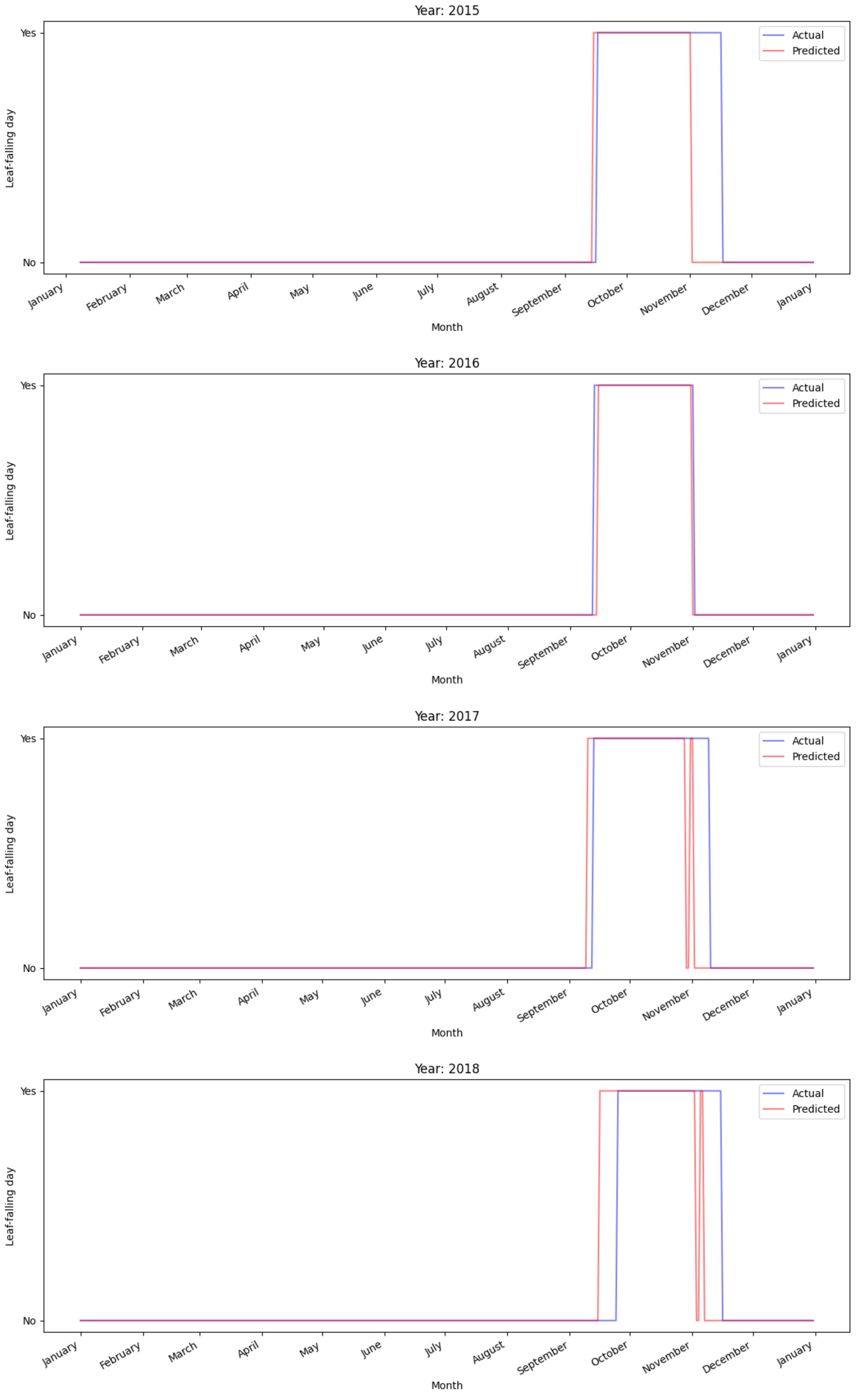}  
\end{subfigure}
\hspace{1cm}  
\begin{subfigure}{.44\textwidth}
  \centering
  \includegraphics[width=\linewidth]{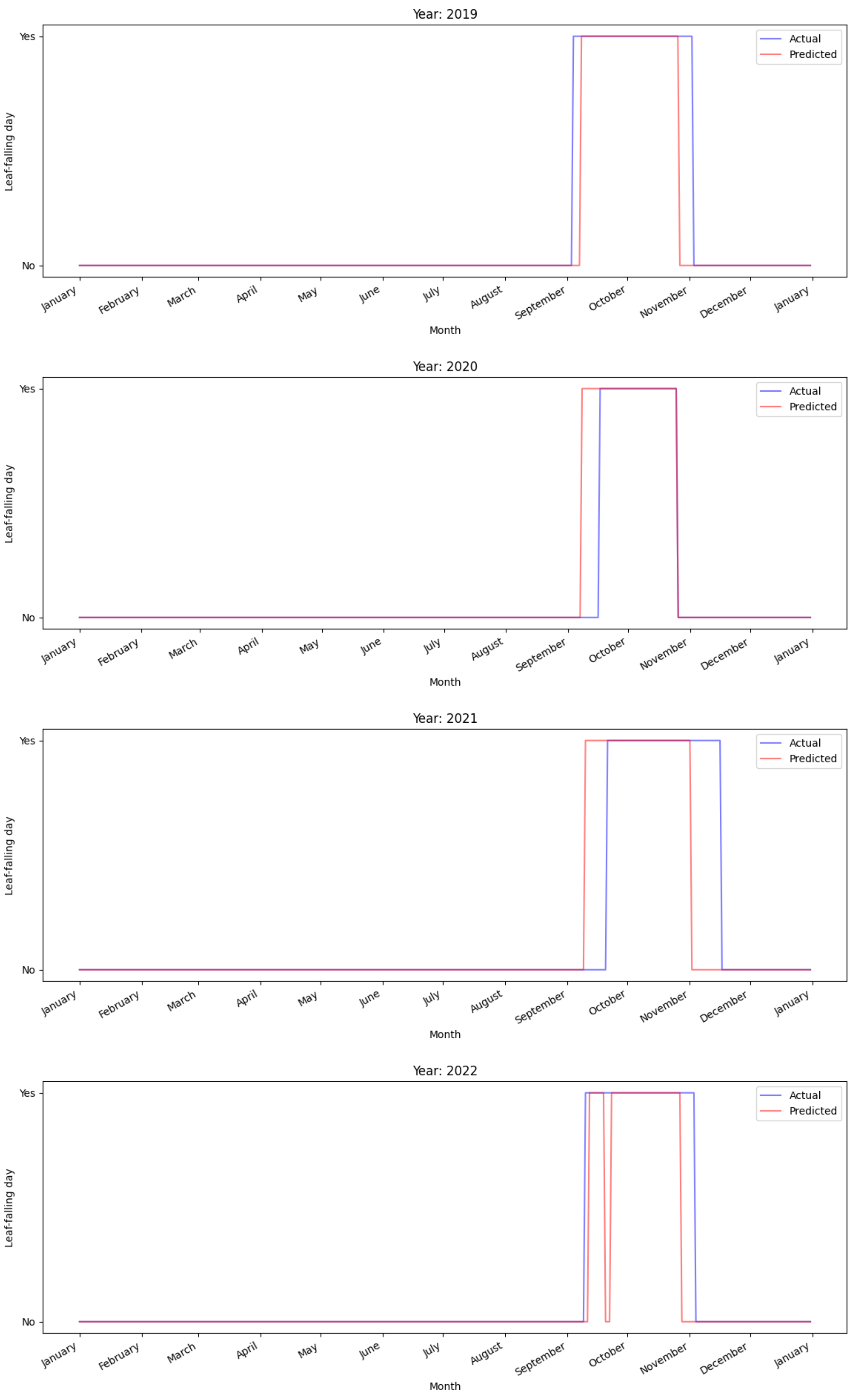}  
\end{subfigure}
\caption{Predicted vs. Actual leaf-falling days per year}
\label{pvsa}
\end{figure}

\section{Leaf-fall trajectory}

\begin{figure}[h]
\centering
\begin{subfigure}{.3\textwidth}
  \centering
  \includegraphics[width=\linewidth]{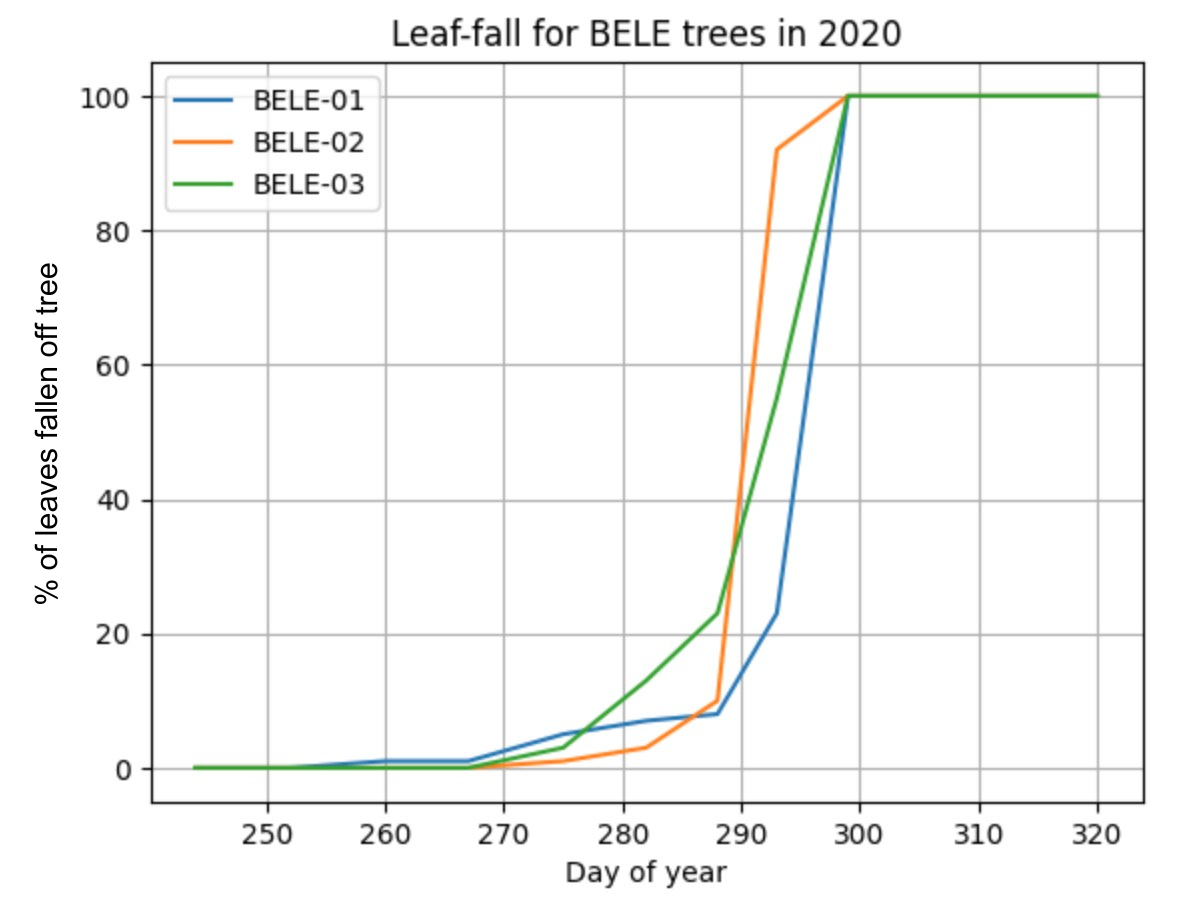}  
  \caption{Betula lenta (BELE)}
  \label{accuracy}
\end{subfigure}
\begin{subfigure}{.3\textwidth}
  \centering
  \includegraphics[width=\linewidth]{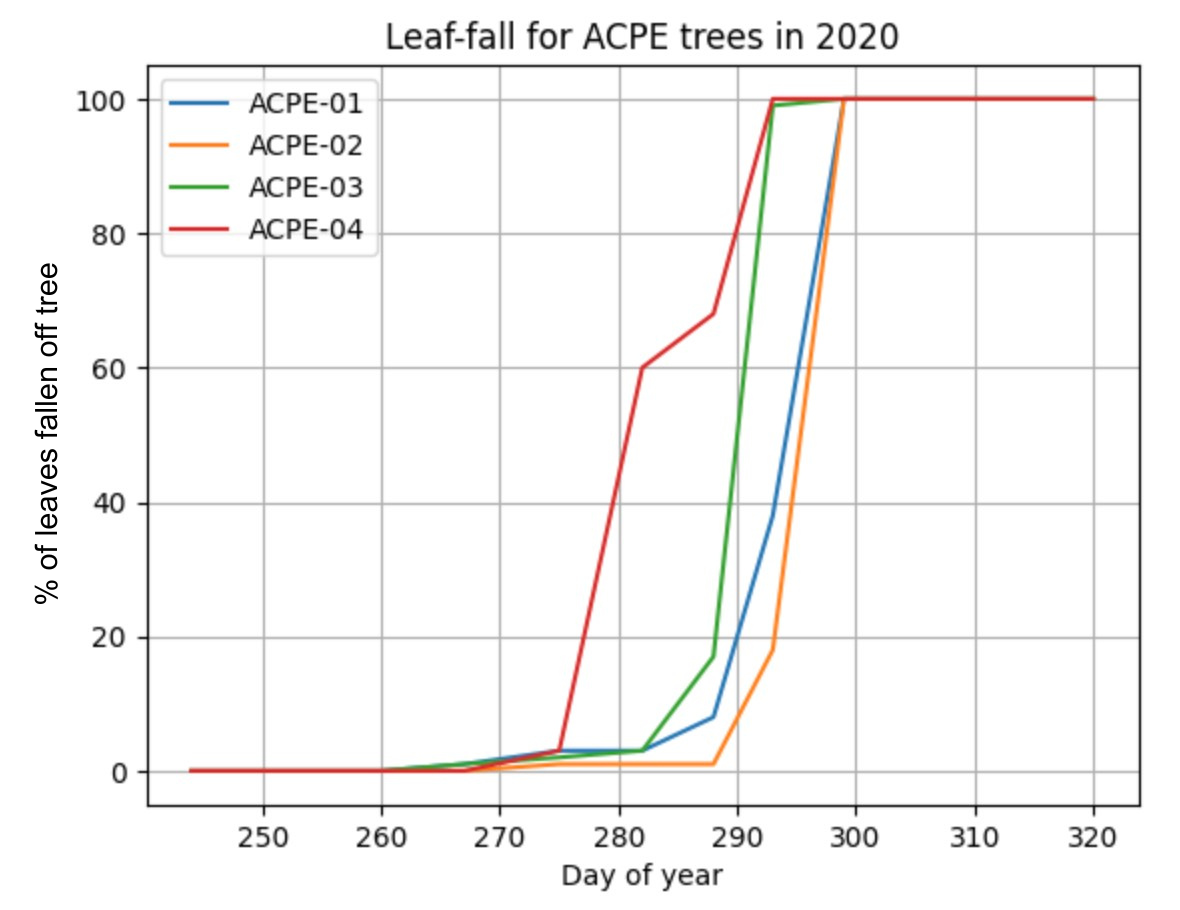}  
  \caption{Acer pensylvanicum (ACPE)}
  \label{loss}
\end{subfigure}
\begin{subfigure}{.3\textwidth}
  \centering
  \includegraphics[width=\linewidth]{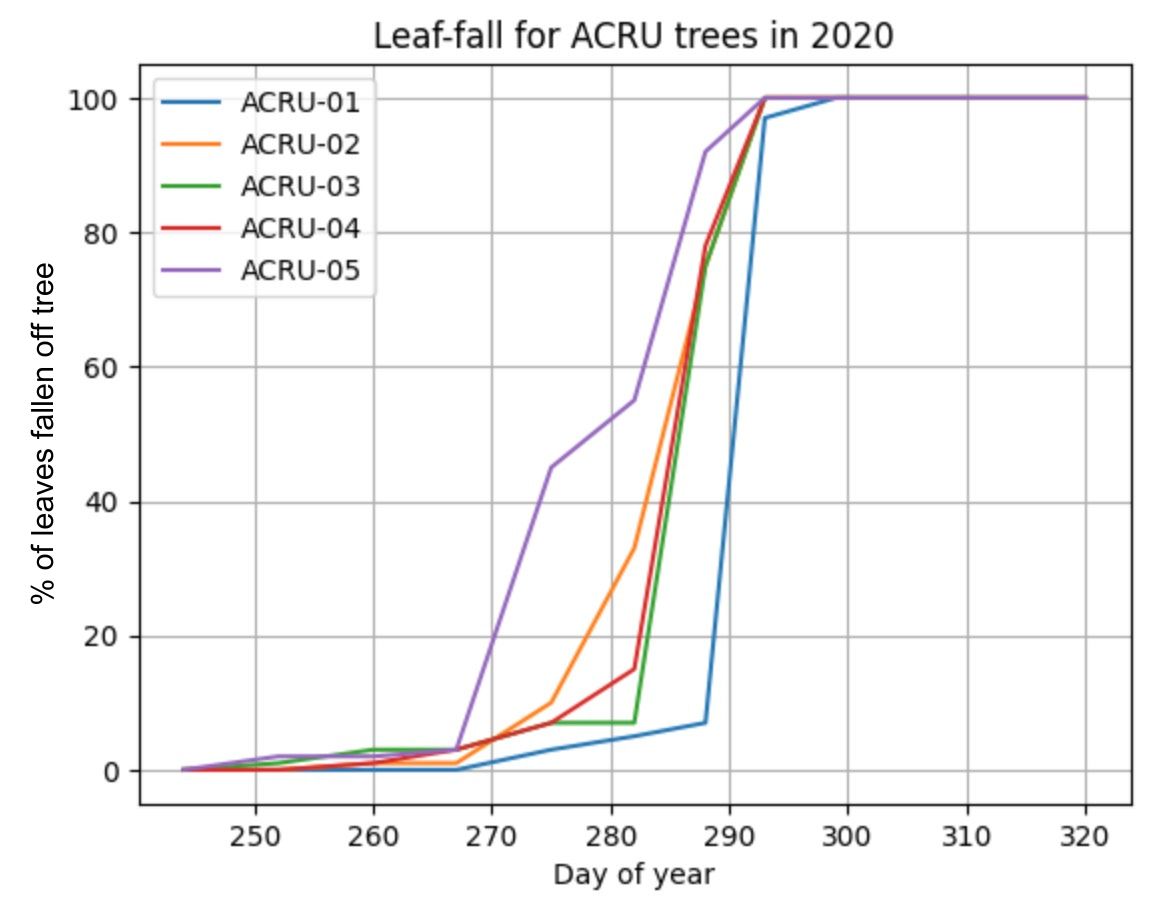}  
  \caption{Acer rubrum (ACRU)}
  \label{loss}
\end{subfigure}
\caption{Leaf-fall for various tree species in Harvard Forest}
\label{accuracy_loss}
\end{figure}

\end{appendices}



\end{document}